\documentclass[10pt,conference,a4paper]{IEEEtran}

\ifCLASSINFOpdf
  \usepackage[pdftex]{graphicx}
\else
\fi
\usepackage{graphbox}
\usepackage[numbers]{natbib}
\usepackage{amsmath,amssymb}
\usepackage{gensymb,nicefrac}
\usepackage{array,multirow}
\usepackage{url}
\usepackage{tabularx}
\usepackage{subcaption,booktabs}
\usepackage{todonotes}
\presetkeys{todonotes}{inline}{}
\usepackage[normalem]{ulem}

\newcommand{\changed}[1]{{#1}}
\newcommand{\removed}[1]{{}}
\begin{document}
\title{Combining GANs and AutoEncoders\\ for Efficient Anomaly Detection}

\author{\IEEEauthorblockN{Fabio Carrara,
Giuseppe Amato,
Luca Brombin,
Fabrizio Falchi,
Claudio Gennaro}
\IEEEauthorblockA{Information Science and Technology Institute (ISTI) of CNR, Pisa, Italy\\
Email: fabio.carrara@isti.cnr.it}}

\newcommand{\MODEL}{{CBiGAN}}

\maketitle

\begin{abstract}
In this work, we propose \MODEL{} --- a novel method for anomaly detection in images, where a consistency constraint is introduced as a regularization term in both the encoder and decoder of a BiGAN. Our model exhibits fairly good modeling power and reconstruction consistency capability.
We evaluate the proposed method on MVTec AD --- a real-world benchmark for unsupervised anomaly detection on high-resolution images --- and compare against standard baselines and state-of-the-art approaches.
Experiments show that the proposed method improves the performance of BiGAN formulations by a large margin and performs comparably to expensive state-of-the-art iterative methods while reducing the computational cost.
We also observe that our model is particularly effective in texture-type anomaly detection, as it sets a new state of the art in this category.
Our code is available at \url{https://github.com/fabiocarrara/cbigan-ad/}.
\end{abstract}

\IEEEpeerreviewmaketitle

\section{Introduction}

\noindent 
In the era of large-scale datasets, the ability to automatically detect outliers (or anomalies) in data \changed{is} relevant \changed{in many application fields}.
\changed{Recognizing anomalous textures is often required e.g. for the detection of defects in industrial manufacturing~\cite{tout2017automatic,bergmann2019mvtec} and large-scale infrastructural maintenance of roads, bridges, rails, etc, while other applications require the detection of more complex shapes and structures, like biomedical applications~\cite{prokopetc2017slim,schlegl2017unsupervised}  dealing for example with early disease detection via medical imaging.}
Even if this problem could be tackled with discriminative approaches --- e.g. as a supervised binary classification problem in which samples can be classified as normal or anomalous --- the cost of collecting a representative training dataset, especially anomalous samples, is often prohibitive.
Thus, interest has grown for \changed{one-class} anomaly detection. %
This is often cast as a learning problem where normal data is modelled exploiting generative approaches, relying on an unlabeled dataset mostly comprised of non-anomalous samples.

Recent developments in deep learning and image representation has introduced visual anomaly detection in many applications as for instance biomedical (e.g. diagnoses aids) and industrial (e.g. quality assurance) applications.
In these applications, classical approaches to anomaly detection, such as statistical-based~\cite{yang2009outlier,cohen2008novelty}, proximity-based~\cite{radovanovic2014reverse,chehreghani2016k}, or clustering-based~\cite{aggarwal2015outlier,manzoor2016fast} approaches, often offer poor performance when applied directly to images, due to the high-dimensionality involved in this type of data, and usually have to rely on the extraction of lower-dimensional hand-crafted or learned features.

Recent approaches directly adopt deep models to contextually map and model the data in a feature space.
In this context, deep generative models for images based on Autoencoders (AEs)~\cite{bergmann2018improving,wang2020advae} or Generative Adversarial Networks (GANs)~\cite{schlegl2017unsupervised,schlegl2019f,akcay2018ganomaly} proved to be effective in anomaly detection.
Most approaches in this category are reconstruction-based: starting from a given sample, they reconstruct the nearest sample on the normal data manifold learned by the model and measure the deviation in a predefined space (e.g. pixel, latent, or combinations) to assess an anomaly.
Autoencoder-based approaches implement this strategy in a straightforward way:
the encoder projects the given sample in the normal data manifold, while the decoder performs the reconstruction.
However, generative models based on autoencoders like VAEs are known to produce blurry reconstructions for photo-realistic images~\cite{dumoulin2017adversarially} and are often outperformed by GANs;
thus in this work, we focus our attention on GAN-based approaches for \changed{one-class} anomaly detection.

\begin{figure}
    \centering
    {\renewcommand{\arraystretch}{0}
    \setlength\tabcolsep{0pt}
    \newcolumntype{C}{>{\centering\arraybackslash}X}
    \begin{tabularx}{\linewidth}{Cccc}
    \centering
    & Query & Reconstruction & Abs. Difference \\[.75ex]
    \rotatebox[origin=c]{90}{Texture - Grid}&%
    \includegraphics[align=c,width=.315\linewidth]{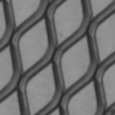}&%
    \includegraphics[align=c,width=.315\linewidth]{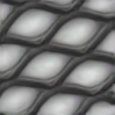}&%
    \includegraphics[align=c,width=.315\linewidth]{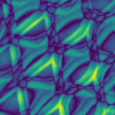} \\
    \rotatebox[origin=c]{90}{Object - Screw}&%
    \includegraphics[align=c,width=.315\linewidth]{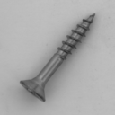}&%
    \includegraphics[align=c,width=.315\linewidth]{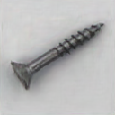}&%
    \includegraphics[align=c,width=.315\linewidth]{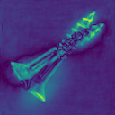} \\
    \end{tabularx}
    }
    \caption{
    \changed{
    Misaligned reconstructions of a standard BiGAN anomaly detector (EGBAD~\cite{zenati2018efficient}) that induce a large distance in pixel space. The consistency loss introduced in \MODEL{} solves this problem.}
    }
    \label{fig:misalign}
\end{figure}

Input reconstruction in GAN-based approaches poses a challenge as, in the standard formulation, there is no encoder that projects the sample in the latent space, and an expensive latent space optimization is often required~\cite{schlegl2017unsupervised}.
Efficient approaches based on Bidirectional GANs (BiGAN) solve this problem by jointly learning an encoder that complements the decoder (i.e. generator) and provide the projection needed (e.g. EGBAD~\cite{zenati2018efficient}). However, reconstructed samples are often misaligned:
the reconstruction goal is only defined by the discriminator, and there is no guarantee that encoding and then decoding a sample will reconstruct it precisely (see Figure~\ref{fig:misalign}).

In this work, we tackle the aforementioned problems and propose a novel method for anomaly detection in images, where we introduce a consistency constraint as a regularization term, on both encoding and decoding parts of a BiGAN. In the reminder of the paper we call our model \MODEL{}.
Our new formulation is able to improve the reconstruction ability with respect to a BiGAN.
It also generalizes both EGBAD~\cite{zenati2018efficient} and AEs by combining the modelling power of the former and the reconstruction consistency of the latter.
We evaluate the proposed method on MVTec AD --- a real-world benchmark for unsupervised anomaly detection on high-resolution images --- and compare against standard baselines and state-of-the-art approaches.
Our results show that the proposed method \changed{solves misalignment problems commonly occurring in GAN formulations improving the performance of BiGAN for complex objects} by a large margin.
\changed{Moreover, our proposal} performs comparably to expensive state-of-the-art iterative methods while reducing computational cost requiring a single-pass evaluation strategy.
We observe that our model is particularly effective on texture-type anomaly detection, as it sets a new state of the art in this category outperforming also models using additional data.

\section{Related Work}

\noindent Most of recent approaches for anomaly detection on images adopt reconstruction-based techniques relying on some sort of deep generative models. %
Autoencoders (standard~\cite{zhou2017anomaly,chen2017outlier,bergmann2018improving} and variational ones~\cite{dehaene2020iterative,wang2020advae}) %
and generative adversarial networks~\cite{schlegl2017unsupervised,akcay2018ganomaly,zenati2018efficient} %
comprise the most commonly adopted techniques:
the former are trained to minimize a reconstruction loss usually in the pixel space, while the latter focus on generating samples indistinguishable from normal data indirectly leading to reconstruction.
Anomaly detection is then implemented by defining a score based on reconstruction error and additional metrics, such as feature matching in latent or intermediate space.
Concerning AE techniques,
\citet{bergmann2018improving} point out that dependencies between pixels are not properly modelled in autoencoders and show that vanilla AEs using a structural similarity index metric (SSIM) loss outperform even complex architectures (e.g. AEs with feature matching and VAEs) that rely on L2 pixel-wise losses.
In \cite{pawlowski2018unsupervised}, the authors instead proposed bayesian convolutional autoencoders: they show that sampling in variational models smooths noisy reconstruction error compared to other AE architectures on medical data.
\citet{zimmerer2019unsupervised} suggest that the gradient w.r.t the input image of the KL-loss in VAE provides useful information on the normality of data and adopt it for anomaly localization.
Building on this concept, \citet{dehaene2020iterative} propose to iteratively integrate these gradients to perform reconstruction by moving samples towards the normality manifold in pixel-space in a similar manner used by adversarial example crafting. %
\citet{golan2018deep} set up a self-supervised multi-classification pretext task of geometric transformation recognition and use the softmax output on transformed samples as feature to characterize abnormality.
Similarly, \citet{huang2019inverse} propose to train a deep architecture to invert geometric transformation of the input and uses the reconstruction error as anomaly score for the input image.

In our work, we build upon GAN-based anomaly detection --- techniques that exploit GANs to learn the anomaly-free distribution of data assuming a uniformly or normally distributed latent space explaining it.
\changed{We focus on the scenario in which we assume a mostly anomaly-free training dataset, and we refer the interested reader to \citet{berg2019unsupervised} for an analysis of GAN-based detectors when this assumption does not hold.}
Among the seminal works in this category, \citet{schlegl2017unsupervised} proposed AnoGAN: the generator is used to perform reconstruction, and the discriminator to extract features of original and reconstructed samples; both are then adopted to define an anomaly score.
The major drawback of this approach is the need of a computational expensive latent space optimization using backpropagation to find the latent space representation that encodes the given sample.
Recent work solve this problem by introducing an encoding module that learns to map samples from the input to the latent space.
\citet{schlegl2019f} propose fast-AnoGAN --- an enhanced AnoGAN with a learned encoding module and adopting the more stable Wasserstein GAN formulation.
\citet{akcay2018ganomaly} propose an adversarially trained encoder-decoder-encoder architecture named GANomaly and define their anomaly score as the L1 error in the latent space between original and reconstructed sample.
\citet{zenati2018efficient} propose EGBAD (Efficient GAN Based Anomaly Detection) that directly adopts a Bidirectional GAN (BiGANs~\cite{donahue2017adversarial})  --- an improved GAN formulation that learns the joint distribution of the latent and input spaces --- and adopt a combination of reconstruction loss and discriminator loss as anomaly score.
In a similar vein, several works propose different architectures that implement some sort of adversarial training~\cite{sabokrou2018avid,venkataramanan2019attention,wang2020advae,zenati2018adversarially}.
\citet{sabokrou2018avid} propose a self-supervised approach for anomaly localization that adopts adversarially-trained fully convolutional network plus a discriminator to detect anomalous pixels. %
\citet{venkataramanan2019attention} propose an adversarially-trained variational autoencoder combined with a specialized attention regularization that encourages the network to model all the parts of the normal input and thus perform precise anomaly localization.

\section{Background}

\newcommand{\R}{\mathbb{R}}
\newcommand{\E}[2]{\mathbb{E}_{#1}\left [ {#2} \right ]}
\newcommand{\x}{\mathbf{x}}
\newcommand{\z}{\mathbf{z}}
\renewcommand{\L}{\mathcal{L}}
\newcommand{\X}{\mathcal{X}}
\newcommand{\Z}{\mathcal{Z}}

\subsection{Generative Adversarial Networks}

\noindent In its basic formulation, a Generative Adversarial Network (GAN~\cite{goodfellow2014generative}) is comprised of a generator $G(\z)$ that generates data starting from a latent variable $\z \sim p(\z)$ and a discriminator $D(\x)$ that discerns whether its input $\x$ is generated by $G$ or coming from the real data distribution $p_\text{data}(\x)$.
$G$ and $D$ are adversarially trained to compete --- $G$ is optimized to fool $D$, while $D$ to tell apart data generated by $G$ and real one.
In a game-theoretic framework, $G$ and $D$ plays the following two-player minimax game
\begin{equation}
    \min_G \max_D \E{\x \sim p_\text{data}(\x)}{\log D(\x)} + \E{\z \sim p(\z)}{\log \left( 1 - D\left (G(\z)\right ) \right) } \,,
\end{equation}
where $D(\x) \in [0, 1]$ indicates how real the discriminator this its input is.
At the Nash equilibrium of the game, $D$ is not able to discern fake samples from real ones, and thus, $G$ is producing samples from $p_\text{data}(\x)$.

\subsection{Wasserstein GAN}
\noindent Reaching the Nash equilibrium is known to be hard due to training instabilities~\cite{salimans2016improved}.
Wasserstein GAN (WGAN~\cite{arjovsky2017wasserstein}) facilitates GAN training by measuring the distance between real and fake data distributions using the Wasserstein distance that assures \changed{more stable} gradients.
In this formulation, $D(\x) \in \R$ is defined as a scalar helper function used to compute the Wasserstein distance that substitutes the value function in the minimax game
\begin{equation} \label{eq:wgan}
    \min_G \max_D \E{\x \sim p_\text{data}(\x)}{D(\x)} - \E{\z \sim p(z)}{D\left( G(\z) \right)} \,.
\end{equation}
The function of $D$ is shifted from a discriminating classifier to a critic that produces authenticity scores and tends to give high scores to real samples and low ones to fake data.
To ensure Lipschitz-continuity of $D$, that is a prerequisite for obtaining Equation~\ref{eq:wgan}, we regularize the norm of the gradient of $D$ with respect to its inputs when optimized as in \cite{gulrajani2017improved}.

\subsection{BiGAN}
\noindent Bidirectional GANs (BiGANs~\cite{donahue2017adversarial}) improves the modeling of the latent space by exposing it to the discriminator together with images generated from it.
An encoder module $E(\x)$ is introduced to map real samples to the corresponding latent space and trained together with $G$.
The discriminator $D(\x, \z) \in [0, 1]$ is trained to discern whether the couple $(\x, \z)$ comes from a real or generated image. 
The minimax problem for the BiGAN is
\begin{equation}
\begin{aligned}
    \min_{G, E} \max_{D} \; &
    \E{\x \sim p_\text{data}(\x)}{\log     D\left(\x,   E(\x)\right) } + \\ &
    \E{\z \sim p(\z)}            {\log\left(1 - D\left(G(\z),\z   \right)\right)} \,.
\end{aligned}
\end{equation}
Thus, fooling $D$ leads $G$ and $E$ to minimize the difference between $(G(\z), \z)$ and $(\x, E(\x))$ couples.
\section{Method}

\begin{figure*}
    \centering
    \includegraphics[width=\linewidth]{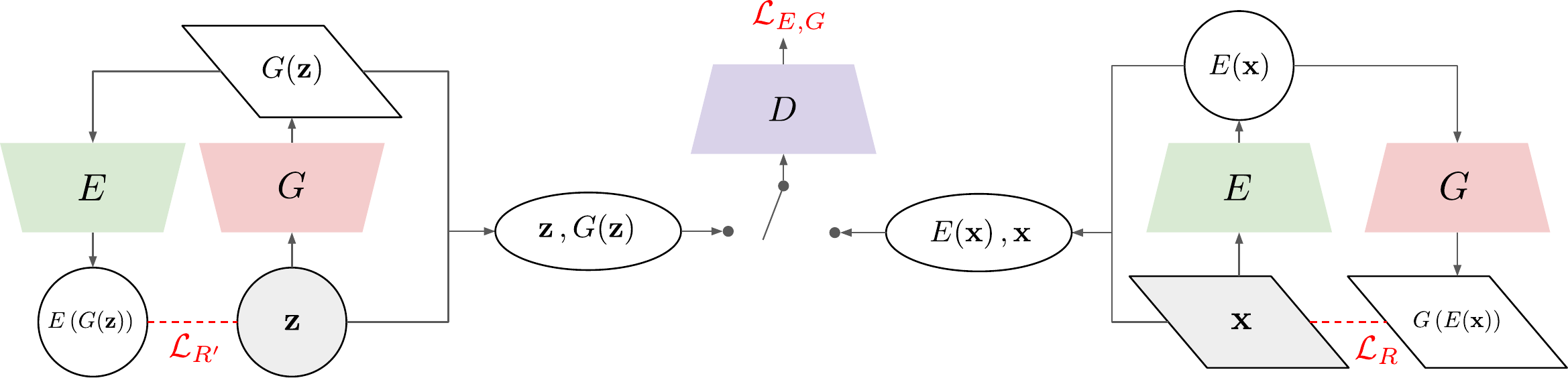}
    \caption{Overview of \MODEL{}.
    Inputs of the model are colored in gray.
    Parallelograms and circles respectively represent samples in the image and latent space, while ellipses represent the inupt of the discrimintor.
    Red dashed lines represent the consistency loss.}
    \label{fig:cbigan}
\end{figure*}

\noindent We tackle anomaly detection as a one-class classification problem --- we assume a training dataset $\{\x_i\} \in \X$ ($\X = \R^{W \times H \times C}$ for images) comprised of only non-anomalous samples.
Given a test sample, we want to label it as normal/non-anomalous/defect-free (we consider it the negative class) or anomalous (positive).
We rely on a GAN-based model to capture the distribution of normal data conditioned to the latent space $\Z = \R^n$.
Similarly to \cite{zenati2018efficient}, we adopt a BiGAN as generative model, but we instantiated it with the Wassestein distance formulation;
its minimax problem is defined as
\begin{equation}
    \min_{G, E} \max_{D} %
    \E{\x \sim p_\text{data}(\x)}{D\left(\x,   E(\x)\right)} - %
    \E{\z \sim p(\z)}            {D\left(G(\z),\z   \right)} \,,
\end{equation}
where $G: \Z \mapsto \X$ is the generator producing fake images from latent representations, $E: \X \mapsto \Z$ is the encoder projecting an image in its corresponding latent representation, and $D: \X \times \Z \mapsto \R$ is the discriminator/critic that scores samples and helps implementing the Wasserstein distance computation.

All three the modules are defined as deep neural networks and are optimized alternatively (once $G$ and $E$, once $D$) with mini-batch gradient descent.
Given a mini-batch of $2N$ elements comprised of $N$ real samples $\{\x_i\}$ and $N$ randomly sampled latent representations $\{\z_i\}$, the losses for the players $\L_{G,E}$ and $\L_D$ are defined as
\begin{equation}
    \L_\changed{E,G} = \frac{1}{N} \sum_{i=1}^N D\left(\x_i, E(\x_i)\right) - \frac{1}{N} \sum_{i=1}^N D\left(G(\z_i), \z_i\right) \,,
\end{equation}
and 
\begin{equation}
    \L_D     =-\frac{1}{N} \sum_{i=1}^N D\left(\x_i, E(\x_i)\right) + \frac{1}{N} \sum_{i=1}^N D\left(G(\z_i), \z_i\right) \,. \label{eq:ld}
\end{equation}
Once trained on normal data, the anomaly detection procedure is reconstruction-based:
given a test sample $\x$, we compute $E(\x)$ to find its closest latent representation on the normal data manifold learned by the model and then compute $G\left(E(\x)\right)$ to build its reconstruction.
Following the AnoGAN approach~\cite{schlegl2017unsupervised}, we define the anomaly score $A(\x)$ as a linear combination of two terms: a) the pixel-based L1 reconstruction error $\L_R(\x)$, and b) the feature-based discriminator error $\L_{f_D}(\x)$. Formally,
\begin{equation}
    A(\x) = (1 - \lambda) \L_{R}(\x) + \lambda \L_{f_D}(\x) \,,
\end{equation}
with
\begin{align}         
    \L_{R}(\x) &= || \x - G\left(E(\x)\right) ||_1 \,, \\
    \L_{f_D}(\x) &= || f_D\left(\x, E(\x)\right) - f_D\left(G\left(E(\x)\right), E(\x)\right) ||_1 \,, \label{eq:ldx}
\end{align}
where {$|| \cdot ||_{1}$} is the {$L_{1}$} norm, $f_D(\x,\z) \in \R^d$ is feature vector extracted from an intermediate output of the discriminator $D$, and $\lambda$ is a balancing hyperparameter whose value is commonly domain-specific.
Intuitively, for an anomalous image both the reconstruction error {$\L_{R}$} and the distance between discriminator features {$\L_{f_D}$} increase, as the encoder $E$ usually maps the input into an out-of-distribution latent representation; therefore, the generator $G$ fails to reconstruct the input, and the discriminator $D$ extracts different representations for the original and reconstructed inputs.

Unfortunately for anomaly detection, the BiGAN training procedure does not put any constraints on $E$ and $G$ being aligned (i.e. $E^{-1} = G$ and vice-versa are not guaranteed,) and this can lead to errors and misalignment in reconstructed samples of normal data and thus to a high false positive rate.
Figure~\ref{fig:misalign} shows an example of this phenomenon;
the reconstructed sample presents a slight rotation with respect to the input sample, and this results in an erroneous high anomaly score.
To cope with this problem, we add a cycle consistency regularization term $\L_C$ to both $E$ and $G$ to promote their alignment.
Formally,
\begin{equation}
\L_C(\x, \z) = \L_R(\x) + \L_{R'}(\z) \,,
\end{equation}
where 
\begin{align}
    \L_{R}(\x) &= || \x - G\left(E(\x)\right) ||_1 \,\text{, and} \\
    \L_{R'}(\z) &= || \z - E\left(G(\z)\right) ||_1 \,.
\end{align}
The new loss for $E$ and $G$ is a linear combination of $\L_{E,G}$ and $\L_C$
\begin{equation}
    \L^*_{E,G} = (1 - \alpha) \L_{E,G} + \alpha \L_C \,,
\end{equation}
where $\alpha$ controls the weight of each contribution.
We refer to the model trained with this formulation as \emph{Consistency BiGAN} (CBiGAN) whose architecture is depicted in Figure~\ref{fig:cbigan}.
Note that setting $\alpha = 1$ we obtain the standard BiGAN approach (EBGAD~\cite{zenati2018efficient}), while with $\alpha = 0$ our model collapses into a non-adversarially-trained double autoencoder (on both latent and input space).

\section{Evaluation}

\subsection{Dataset}
\noindent We tested and compared our approach on MVTec AD~\cite{bergmann2019mvtec} --- a recent real-world benchmark specifically tailored for unsupervised anomaly detection that is increasingly adopted in recent literature on this topic~\cite{venkataramanan2019attention,huang2019inverse,liu2019towards}.
It is comprised of over 5,000 high-resolution industrial images divided into 10 objects and 5 textures categories;
for each category, MVTec AD provides a training set of non-anomalous (defect-free) images and a labelled test set containing both anomalous and normal images.
Details of MVTec AD are reported in Table~\ref{tab:mvtec-ad:details}, and examples (together with the reconstruction performed by our model) are depicted in Figure~\ref{fig:grid}.

\begin{table}
\newcolumntype{C}{>{\centering\arraybackslash}X}

    \centering
    \begin{tabular}{cl*{4}{c}}
        \toprule
    &               & Train & \multicolumn{2}{c}{Test} & Image Side \\ \cmidrule(lr){4-5}
    &               &     N &        N  &          P   &            \\ \midrule
    \parbox[t]{2mm}{\multirow{5}{*}{\rotatebox[origin=c]{90}{Textures}}}
    &   Carpet      &   280 &       28  &         89   &       1024 \\
    &   Grid        &   264 &       21  &         57   &       1024 \\
    &   Leather     &   245 &       32  &         92   &       1024 \\
    &   Tile        &   230 &       33  &         84   &        840 \\
    &   Wood        &   247 &       19  &         60   &       1024 \\ \midrule
    \parbox[t]{2mm}{\multirow{10}{*}{\rotatebox[origin=c]{90}{Objects}}}
    &   Bottle      &   209 &       20  &         63   &        900 \\
    &   Cable       &   224 &       58  &         92   &       1024 \\
    &   Capsule     &   219 &       23  &        109   &       1000 \\
    &   Hazelnut    &   391 &       40  &         70   &       1024 \\
    &   Metal nut   &   220 &       22  &         93   &        700 \\
    &   Pill        &   267 &       26  &        141   &        800 \\
    &   Screw       &   320 &       41  &        119   &       1024 \\
    &   Toothbrush  &    60 &       12  &         30   &       1024 \\
    &   Transistor  &   213 &       60  &         40   &       1024 \\
    &   Zipper      &   240 &       32  &        119   &       1024 \\ \bottomrule
    \end{tabular}
    \caption{MVTec-AD dataset details. N = \# of negative (anomaly-free) samples. P = \# of positive (anomalous) samples. All images are squared with side length given in the last column.}
    \label{tab:mvtec-ad:details}

\end{table}

\subsection{Implementation Details}

\begin{figure*}[t]
    \centering
    \includegraphics[width=\textwidth]{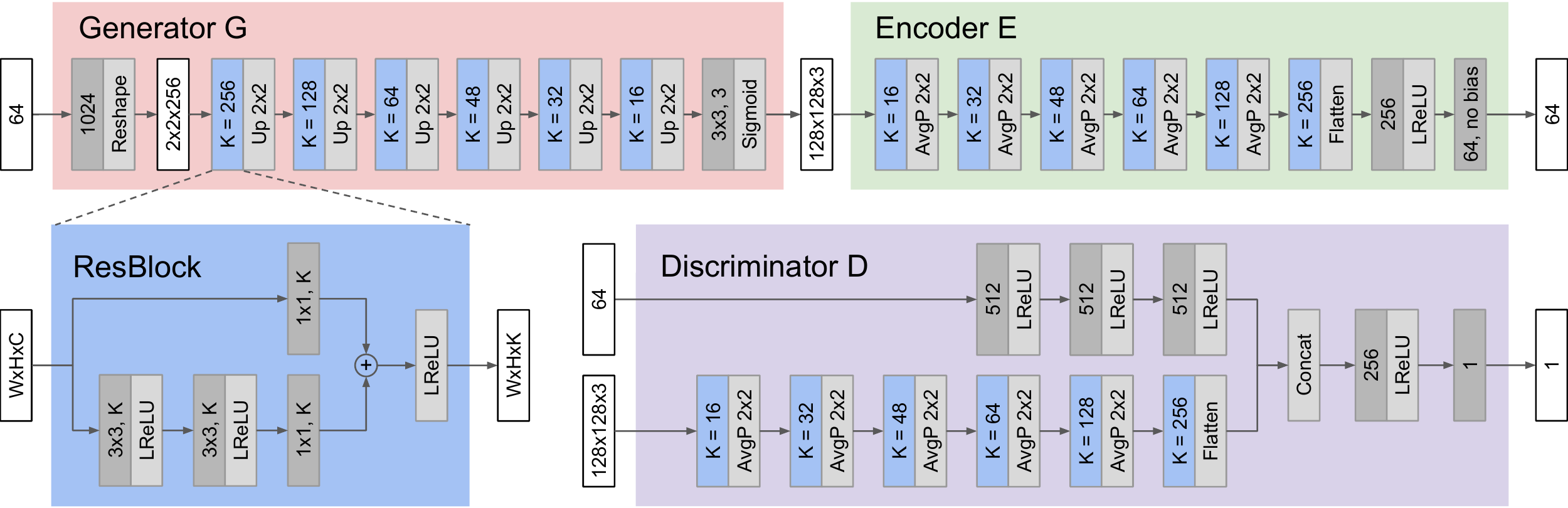}
    \caption{Our implementation of $E$, $G$, and $D$ of \MODEL{} for MVTec AD. White blocks specify input and output shapes, and dark gray blocks indicate learnable convolutions (\texttt{<filter\_width}\textsf{ x }\texttt{filter\_height, num\_filters>}) or fully-connected layers (\texttt{<num\_outputs>}).
    The negative slope of LeakyReLU activations is $0.2$.
    Upsampling adopts bilinear interpolation.}
    \label{fig:archs}
\end{figure*}

\noindent We follow the evaluation procedure commonly adopted in related work~\cite{schlegl2017unsupervised} in the unsupervised setting.
We implemented $E$, $G$, and $D$ as standard residual convolutional networks.
Architectural details are reported in Figure~\ref{fig:archs}.
We train %
\MODEL{} separately for each category using the corresponding anomaly-free training set.
Different preprocessing are adopted depending on the type of object considered:
\begin{itemize}
    \item for `Object' categories, we resize the input images to 128x128, and we apply data augmentation in the form of random rotation in the $[-45\degree, +45\degree]$ range whenever the orientation of the object is independent of its abnormality --- for Bottle, Hazelnut, Metal nut, and Screw categories;
    \item for `Texture' categories, we resize the input image to 512x512 and processed in 64x64 patches;
    when training, we randomly crop 64x64 patches from the input image and apply random clockwise rotation in the $[0\degree, +45\degree]$ range, and 
    during testing, we divide the input image in 64x64 patches, derive a local anomaly score for each patch, and obtain a global score by picking the maximum local score.
\end{itemize}
For all models and categories, we use a 64-dimensional latent space ($\Z = \R^{64}$) sampled using a normal distribution.
We tuned the weight of the consistency loss $\alpha$  experimentally; good values are $\alpha = 10^{-4}$ for `Objects' and $\alpha \in [10^{-4}, 10^{-5}]$ for `Textures' depending on the importance of pixel-level details in detecting an anomaly.
Note that small values of $\alpha$ are needed to balance the usually large values of the consistency loss term $\L_{C}$ that depends on the input and latent dimensions.
We also produce the results for EGBAD~\cite{zenati2018efficient} by setting $\alpha = 0$.
All models are trained using the Adam optimizer with a learning rate of $1e^{-4}$. %
Following \cite{schlegl2017unsupervised}, we set $\lambda = 0.1$ when computing the anomaly score $A(\x)$ for testing images.
    
\subsection{Evaluation Metrics}
\noindent To quantitatively evaluate the quality of the tested approaches, for each category we compute
\begin{itemize}
    \item the maximum Balanced Accuracy $B=\frac{\text{TPR} + \text{TNR}}{2}$ obtained when varying the threshold on the anomaly score;
    this is often reported by previous work~\cite{bergmann2019mvtec,venkataramanan2019attention} and referred to as ``the mean of the ratio of correctly classified samples of anomaly-free (TNR) and anomalous images (TPR)''.
    The choice of the threshold is equivalent to the maximization of the Youden's index~\cite{sokolova2006beyond} $J = \text{TPR} + \text{TNR} - 1$.
    \item the Area Under the ROC Curve (auROC), commonly adopted as a threshold-independent quality metric for classification.
\end{itemize}
We also report per-type and overall means of the above metrics as a unique indicator of the quality of the compared methods.

\subsection{Comparison with State of the Art}
\noindent We compare our approaches with several methods tackling \changed{one-class} anomaly detection and obtaining state-of-the-art results on MVTec AD.
We %
divide the methods into
\begin{itemize}
\item
\emph{iterative methods} that require an iterative optimization for each sample on which anomaly detection is performed and include AnoGAN~\cite{schlegl2017unsupervised} and VAE-grad~\cite{dehaene2020iterative}, and
\item
\emph{single-pass methods} that perform anomaly detection in the testing phase with a single forward evaluation of the model and include AE$_\text{L2}$ and AE$_\text{SSIM}$~\cite{bergmann2018improving}, AVID~\cite{sabokrou2018avid}, LSA~\cite{abati2019latent}, EGBAD~\cite{zenati2018efficient}, GeoTrans~\cite{golan2018deep}, GANomaly~\cite{akcay2018ganomaly}, ITAE~\cite{huang2019inverse}, and our \MODEL{}.
\end{itemize}
In the latter, we also include CAVGA~\cite{venkataramanan2019attention} even if the authors report results using models trained with additional data and thus not directly comparable with the other methods.
For EGBAD, results are coming from our implementation, as it is generalized by \MODEL{} and can be obtained with $\alpha=0$.

\subsection{Results}
\noindent Tables~\ref{tab:mvtec-ad:bal-acc} and~\ref{tab:mvtec-ad:auroc} report results for each compared methods and for each category of MVTec AD.
Compared to EGBAD, that is the approach we extend, we observe that the consistency constraint added in our model consistently improve the detection of anomalies in all the categories (with Zipper being the only exception) by at most 49\%, achieving a +15\% improvement on the overall balanced accuracy; the same conclusion can be drawn also from the auROC metric.
\changed{These performance gains are achieved maintaining the exact same computational cost of EGBAD during prediction (a single forward pass of $E$ and $G$) and avoiding expensive iterative methods. The only additional cost to the BiGAN approach is the computation of the consistency loss $\L_C$ during the offline training phase.}

\changed{Figure~\ref{fig:samples} reports some relevant examples of the differences between EGBAD and \MODEL{}.
Note that our higher quality of reconstructions is due to the recovered alignment and the higher color fidelity in background colors, that overall inducing smaller differences in non-anomalous areas.
The Zipper class is challenging for both EGBAD and our model, as we deem the anomalous parts of the image (usually dents) are small with respect to other source of variability (the border of the zipper) that need to be modeled (see Figure~\ref{fig:samples} last row); thus both models tend to have a high reconstruction loss for both normal and anomalous samples.
}

Both metrics also show that \MODEL{} improves on all the compared methods when dealing with textures anomalies, reaching respectively a mean balanced accuracy and mean auROC of 0.84 and 0.85 and outperforming also methods using additional data.
When all categories are concerned, our method performs comparably to both other single-pass and iterative methods in terms of overall balanced accuracy, respectively obtaining a +3\% (vs AVID, LSA) and -1\% (vs VAE-grad) performance with respect to the second best methods.

Experiments on objects display a more complex situation; among single-pass methods, an absolute best does not emerge, and different categories benefit from specific peculiarities of each method.
\MODEL{} does improve on vanilla BiGAN (EGBAD) on objects, but overall, our method offers an average performance comparable or slightly degraded ($\sim$2-3\%) with respect to other single-pass methods.
Note that the performance of the vanilla L2 autoencoder suggests that tuning the $\alpha$ hyperparameter for each particular object category could further increase performance of \MODEL{} \changed{(e.g. Zipper could benefit from a higher $\alpha$), but for sake of simplicity, we refrain from exploring class-specific parameter values in this work and prefer presenting results for fixed reasonable values}.
Among the few methods reporting the auROC on MVTec AD data, our model still achieves state-of-the-art performance on textures and offers a better or comparable performance with respect to other methods on objects with the only exception represented by ITAE~\cite{huang2019inverse} --- a data-augmentation-based denoising autoencoder.
At the time of writing, we abstain from discussing the performance of ITAE due to reproducibility issues.

\begin{figure}
    \centering
    {
    \renewcommand{\arraystretch}{0}
    \setlength\tabcolsep{0pt}
    \newcolumntype{C}{>{\centering\arraybackslash}X}
    \begin{tabularx}{\linewidth}{Cccc}
    & Query & Recon. & Diff. \\[.75ex]
    \rotatebox[origin=c]{90}{\small Texture - Grid}&%
    \includegraphics[align=c,width=.315\linewidth]{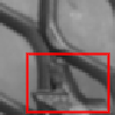}&%
    \includegraphics[align=c,width=.315\linewidth]{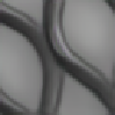}&%
    \includegraphics[align=c,width=.315\linewidth]{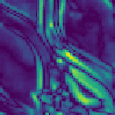}\\

    \rotatebox[origin=c]{90}{\small Obj. - Toothbrush}&%
    \includegraphics[align=c,width=.315\linewidth]{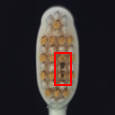}&%
    \includegraphics[align=c,width=.315\linewidth]{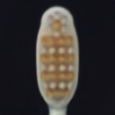}&%
    \includegraphics[align=c,width=.315\linewidth]{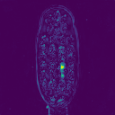}\\
    \end{tabularx}
    }
    \caption{
    \changed{
    Examples of anomalous images from MVTec-AD (1$^\text{st}$ col.), the reconstructions of our \MODEL{} (2$^\text{nd}$ col.), and visualization of the pixel-wise absolute difference (3$^\text{rd}$ col.).}}
    \label{fig:grid}
\end{figure}

\renewcommand{\b}{\bfseries}
\renewcommand{\u}[1]{\underline{#1}}
\newcommand{\m}[1]{{\scriptsize #1}}
\newcolumntype{C}{>{\centering\arraybackslash}X}
\newcolumntype{M}[1]{>{\centering\arraybackslash}m{#1}}

\begin{table*}
    \setlength{\tabcolsep}{2.5pt}
    \centering
    \begin{tabularx}{\linewidth}{X*{5}{c}|c*{10}{c}|cM{1cm}}
        \toprule
                & \multicolumn{6}{c}{Textures}                                             & \multicolumn{11}{c}{Objects}                              & \multirow{2}{1cm}[-.87ex]{\centering Overall\\Mean} \\
                \cmidrule{2-7} \cmidrule(l){8-18}
                & \m{Carpet} & \m{Grid} & \m{Leather} & \m{Tile} & \m{Wood} & \m{\it Mean} & \m{Bottle} & \m{Cable} & \m{Capsule} & \m{Hazelnut} & \m{MetalNut} & \m{Pill} & \m{Screw} & \m{Toothbrush} & \m{Transistor} & \m{Zipper} & \m{\it Mean} &     \\
\cmidrule{2-7} \cmidrule(l){8-18} \cmidrule(lr){19-19}
\multicolumn{5}{l}{\emph{Iterative methods}} \\
\cmidrule(r){1-1} %
AnoGAN~\cite{schlegl2017unsupervised}$^\dagger$           &     0.49 &     0.51 &     0.52 &     0.51 &     0.68 &\it     0.54 &     0.69 &     0.53 &    0.58 &    0.50 &     0.50 &    0.62 &    0.35 &     0.57 &    0.67 &    0.59 &\it    0.56 &    0.55 \\
VAE-grad~\cite{dehaene2020iterative}$^\dagger$            &     0.67 &     0.83 &     0.71 &     0.81 &\b\u{0.89}&\it     0.78 &     0.86 &     0.56 &\b  0.86 &    0.74 &\b\u{0.78}&    0.80 &    0.71 &     0.89 &    0.70 &    0.67 &\it\b  0.76 & \b 0.77 \\
\\
\multicolumn{5}{l}{\emph{Single-pass methods}} \\
\cmidrule(r){1-1} %
AE$_\text{SSIM}$~\cite{bergmann2018improving}$^\star$     &     0.67 &     0.69 &     0.46 &     0.52 &     0.83 &\it     0.63 & \b  0.88 &     0.61 &    0.61 &    0.54 &     0.54 &    0.60 &    0.51 &     0.74 &    0.52 &    0.80 &\it    0.64 &    0.63 \\
AE$_\text{L2}$~\cite{bergmann2018improving}$^\star$       &     0.50 &     0.78 &     0.44 &     0.77 &     0.74 &\it     0.65 &     0.80 &     0.56 &    0.62 & \b 0.88 &     0.73 &    0.62 &    0.69 &\b\u{0.98}&    0.71 &    0.80 &\it    0.74 &    0.71 \\
AVID~\cite{sabokrou2018avid}$^\dagger$                    &     0.70 &     0.59 &     0.58 &     0.66 &     0.83 &\it     0.67 & \b  0.88 &     0.64 &    0.85 &    0.86 &     0.63 &\b  0.86 &    0.66 &     0.73 &    0.58 &    0.84 &\it    0.75 &    0.73 \\
LSA~\cite{abati2019latent}$^\dagger$                      &\b   0.74 &     0.54 &     0.70 &     0.70 &     0.75 &\it     0.69 &     0.86 &     0.61 &    0.71 &    0.80 &     0.67 &    0.85 &\b  0.75 &     0.89 &    0.50 &\b  0.88 &\it    0.75 &    0.73 \\
EGBAD~\cite{zenati2018efficient}                          &     0.60 &     0.50 &     0.65 &     0.73 &     0.80 &\it     0.66 &     0.68 &     0.66 &    0.55 &    0.50 &     0.55 &    0.63 &    0.50 &     0.48 &    0.68 &    0.59 &\it    0.58 &    0.61 \\
\MODEL{} (ours)                                           &     0.60 &\b\u{0.99}&\b\u{0.87}&\b\u{0.84}&     0.88 &\it\b\u{0.84}&     0.84 &\b\u{0.73}&    0.58 &    0.75 &     0.67 &    0.76 &    0.67 &     0.97 &\b  0.74 &    0.55 &\it    0.73 &    0.76 \\
\\
\multicolumn{5}{l}{\emph{Methods using additional data}} \\
\cmidrule(r){1-1} %
CAVGA-D$_u$~\cite{venkataramanan2019attention}            &     0.73 &     0.75 &     0.71 &     0.70 &     0.85 &\it     0.75 &     0.89 &     0.63 &    0.83 &    0.84 &     0.67 &    0.88 &    0.77 &     0.91 &    0.73 &    0.87 &\it    0.80 &    0.78 \\
CAVGA-R$_u$~\cite{venkataramanan2019attention}            &  \u{0.78}&     0.78 &     0.75 &     0.72 &     0.88 &\it     0.78 &  \u{0.91}&     0.67 & \u{0.87}&    0.87 &     0.71 & \u{0.91}& \u{0.78}&     0.97 & \u{0.75}& \u{0.94}&\it \u{0.84}& \u{0.82}\\ 
        \bottomrule
    \end{tabularx}
    \caption{MVTec-AD: Maximum Balanced Accuracy $ B = \nicefrac{\left (\text{TPR} + \text{TNR}\right)}{2}$ when varying the anomaly score threshold for each category.
    The last column reports the average balanced accuracy.
    $^\star$Results from \cite{bergmann2019mvtec}.
    $^\dagger$Results from \cite{venkataramanan2019attention}.
    Results in \textbf{boldface} indicate best performing methods using only the training data provided by MVTec-AD, while \underline{underlined} results indicate the best method also among the ones using additional data.
    }
    \label{tab:mvtec-ad:bal-acc}
\end{table*}

\begin{table*}
    \setlength{\tabcolsep}{2.5pt}
    \centering
    \begin{tabularx}{\linewidth}{X*{5}{c}|c*{10}{c}|cM{1cm}}
        \toprule
                & \multicolumn{6}{c}{Textures}                                             & \multicolumn{11}{c}{Objects}                              & \multirow{2}{1cm}[-.87ex]{\centering Overall\\Mean} \\
                \cmidrule{2-7} \cmidrule(l){8-18}
                & \m{Carpet} & \m{Grid} & \m{Leather} & \m{Tile} & \m{Wood} & \m{\it Mean} & \m{Bottle} & \m{Cable} & \m{Capsule} & \m{Hazelnut} & \m{MetalNut} & \m{Pill} & \m{Screw} & \m{Toothbrush} & \m{Transistor} & \m{Zipper} & \m{\it Mean} &     \\

\cmidrule{2-7} \cmidrule(l){8-18} \cmidrule(lr){19-19}
AE$_\text{L2}^\dagger$                     &    0.64 &    0.83 &    0.80 &    0.74 & \b 0.97 & \it   0.80 &    0.65 &    0.64 &    0.62 &    0.73 &    0.64 &    0.77 & \b 1.00 &    0.77 &    0.65 &    0.87 & \it   0.74 &    0.75 \\
GeoTrans~\cite{golan2018deep}$^\dagger$    &    0.44 &    0.62 &    0.84 &    0.42 &    0.61 & \it   0.59 &    0.74 &    0.78 &    0.67 &    0.36 & \b 0.81 &    0.63 &    0.50 &    0.97 & \b 0.87 &    0.82 & \it   0.71 &    0.67 \\
GANomaly~\cite{akcay2018ganomaly}$^\dagger$&    0.70 &    0.71 &    0.84 &    0.79 &    0.83 & \it   0.77 &    0.89 &    0.76 & \b 0.73 &    0.79 &    0.70 &    0.74 &    0.75 &    0.65 &    0.79 &    0.75 & \it   0.76 &    0.76 \\
ITAE~\cite{huang2019inverse}$^\dagger$     & \b 0.71 &    0.88 & \b 0.86 &    0.74 &    0.92 & \it   0.82 & \b 0.94 & \b 0.83 &    0.68 & \b 0.86 &    0.67 &    0.79 & \b 1.00 & \b 1.00 &    0.84 & \b 0.88 & \it\b 0.85 & \b 0.84 \\
EGBAD~\cite{zenati2018efficient}           &    0.52 &    0.54 &    0.55 &    0.79 &    0.91 & \it   0.66 &    0.63 &    0.68 &    0.52 &    0.43 &    0.47 &    0.57 &    0.46 &    0.64 &    0.73 &    0.58 & \it   0.57 &    0.60 \\
\MODEL{} (ours)                            &    0.55 & \b 0.99 &    0.83 & \b 0.91 &    0.95 & \it\b 0.85 &    0.87 &    0.81 &    0.56 &    0.77 &    0.63 & \b 0.81 &    0.58 &    0.94 &    0.77 &    0.53 & \it   0.73 &    0.77 \\
        \bottomrule
    \end{tabularx}
    \caption{MVTec-AD: Area Under the ROC curve (auROC). $^\dagger$Results from~\cite{huang2019inverse}}
    \label{tab:mvtec-ad:auroc}
\end{table*}

\begin{figure}[t]
{
\renewcommand{\arraystretch}{0}
\newcommand{\sampleRow}[1]{%
\includegraphics[width=.2\linewidth]{img/samples/#1_q.png}&%
\includegraphics[width=.2\linewidth]{img/samples/#1_gen_e.png}&%
\includegraphics[width=.2\linewidth]{img/samples/#1_diff_e.png}&%
\includegraphics[width=.2\linewidth]{img/samples/#1_gen_c.png}&%
\includegraphics[width=.2\linewidth]{img/samples/#1_diff_c.png}\\%
}

\setlength\tabcolsep{0pt}
\centering
\begin{tabular}{ccccc}
\centering
& \multicolumn{2}{c}{EGBAD~\cite{zenati2018efficient}} & \multicolumn{2}{c}{\MODEL{} (ours)} \\[.75ex] \cmidrule(lr){2-3} \cmidrule(lr){4-5}
Query & Recon. & Diff. & Recon. & Diff. \\[.75ex]
\sampleRow{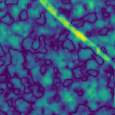}
\sampleRow{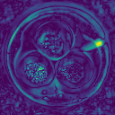}
\sampleRow{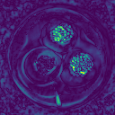}
\sampleRow{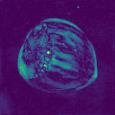}
\sampleRow{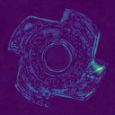}
\sampleRow{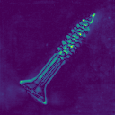}
\sampleRow{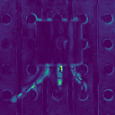}
\sampleRow{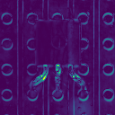}
\sampleRow{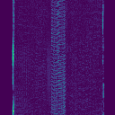}
\end{tabular}
}
\caption{
\changed{\MODEL{} vs EGBAD --- 
input images (1$^\text{st}$ col., red boxes indicate anomalies), reconstructions (2$^\text{nd}$ \& 4$^\text{th}$ col.), and per-pixel absolute differences (3$^\text{rd}$ \& 5$^\text{th}$ col.).
Best viewed in electronic format.
}}
\label{fig:samples}
\end{figure}

\section{Conclusion}
\noindent We tackled \changed{one-class} anomaly detection of images using deep generative models and reconstruction-based approaches.
We proposed \MODEL{} --- an improved Bidirectional GAN model with a consistency regularization on both the encoder and decoder modules.
Our model generalizes and combines both BiGANs and Autoencoders to retain the modelling power of the former and the reconstruction accuracy of the latter.
We evaluated our proposal on MVTec AD --- a real-world benchmark for unsupervised visual anomaly detection with focus on industrial applications.
The results of our experiments showed that our proposal greatly improves the reconstruction ability (and thus performance) of vanilla bidirectional GANs \changed{on both texture and object categories} while maintaining its efficiency at test time.
Our \MODEL{} is particularly effective on texture-type images where it sets the new state of the art obtaining the best mean accuracy and auROC among competing methods including expensive iterative ones and approaches using additional data.
Concerning object-type images, we observed that no particular method prevails on others, as each object category comes with different peculiarities.
In this context, our model provides an average performance comparable with other efficient (single-pass) methods.
In future work, we plan to evaluate our model also on the task of anomaly localization and gain insight on the effect of $\alpha$ in that context. 

\section*{Acknowledgment}

\noindent This work was partially funded by ``Automatic Data and documents Analysis to enhance human-based processes" (ADA, CUP CIPE D55F17000290009) and the AI4EU project (funded by the EC, H2020 - Contract n. 825619). We gratefully acknowledge the support of NVIDIA Corporation with the donation of a Tesla K40 GPU and a Jetson TX2 used for this research.

\bibliographystyle{IEEEtranN}
\bibliography{biblio}

\end{document}